\documentclass{article}

\usepackage[final, nonatbib]{neurips_2019}

\usepackage[utf8]{inputenc}     % allow utf-8 input
\usepackage[T1]{fontenc}        % use 8-bit T1 fonts
\usepackage{hyperref}           % hyperlinks
\usepackage{url}                % simple URL typesetting
\usepackage{booktabs}           % professional-quality tables
\usepackage{amsfonts}           % blackboard math symbols
\usepackage{microtype}          % microtypography
\usepackage{graphicx}           % figures

\title{Deep Hyperedges: a Framework for Transductive and Inductive Learning on Hypergraphs}

\author{%
  Josh Payne \\
  IBM T. J. Watson Research Center, Yorktown Heights, NY 10598 \\
  Stanford University, Stanford, CA 94305 \\
  \texttt{jfp@cs.stanford.edu}
}

\begin{document}

\maketitle

\begin{abstract}
From social networks to protein complexes to disease genomes to visual data, hypergraphs are everywhere. However, the scope of research studying deep learning on hypergraphs is still quite sparse and nascent, as there has not yet existed an effective, unified framework for using hyperedge and vertex embeddings jointly in the hypergraph context, despite a large body of prior work that has shown the utility of deep learning over graphs and sets. Building upon these recent advances, we propose \textit{Deep Hyperedges} (DHE), a modular framework that jointly uses contextual and permutation-invariant vertex membership properties of hyperedges in hypergraphs to perform classification and regression in transductive and inductive learning settings. In our experiments, we use a novel random walk procedure and show that our model achieves and, in most cases, surpasses state-of-the-art performance on benchmark datasets. Additionally, we study our framework's performance on a variety of diverse, non-standard hypergraph datasets and propose several avenues of future work to further enhance DHE.
\end{abstract}

\section{Introduction}
As data becomes more plentiful, we find ourselves with access to increasingly complex networks that can be used to express many different types of information. Hypergraphs have been used to recommend music \cite{music}, model cellular and protein-protein interaction networks \cite{cells, hypergraphlet}, classify images and perform 3D object recognition \cite{images, gao}, diagnose Alzheimer's disease \cite{alzheimers}, analyze social relationships \cite{social, hgnn}, and classify gene expression \cite{arrayCGH}---the list goes on. However, despite the clear utility of hypergraphs as expressive objects, the field of research studying the application of deep learning to hypergraphs is still emerging. Certainly, there is a much larger and more mature body of research studying the application of deep learning to graphs, and the first notions of graph neural networks were introduced merely just over a decade ago by Gori et al. in \cite{gori} and Scarselli et al. in \cite{gnn}. But despite its nascency, the field of deep learning on hypergraphs has seen a rise in attention from researchers recently, which we investigate in Section \ref{related}. 

In 2017, Zaheer et al. \cite{deepsets} proposed a novel architecture, \textit{DeepSets}, for performing deep machine learning tasks on permutation-invariant and equivariant objects. This work was extended and generalized by Hartford et al. \cite{multisets} to describe and study interactions between multiple sets. 

In this work, we go a step further, proposing a novel framework that utilizes ideas from context-based graph embedding approaches and permutation-invariant learning to perform transductive and inductive inference on hypergraphs---sets of sets with underlying contextual graph structure. This framework can be used for classification and regression of vertices and hyperedges alike; this work focuses on the classification of hyperedges.

\section{Related Work}
\label{related}
In 2013, Mikolov et al. proposed an unsupervised learning procedure, \textit{skip-gram}, in \cite{word2vec} which uses negative sampling to create context-based embeddings for terms given sentences in a corpus of text. This procedure was used by Perozzi et al. in \textit{DeepWalk} \cite{deepwalk} to treat random walks on a graph as ``sentences" and vertices as ``terms", which outperformed existing spectral clustering and weighted vote-based relational neighbor classifiers \cite{spectral, ane, tang3,latent,rel}. Similar random walk-based approaches followed, such as random walks with bias parameters \cite{node2vec}, methods that utilize network attributes as additional features \cite{planetoid, DANE, tadw} and approaches that could be extended to perform inductive learning \cite{planetoid, graphsage}. Graph convolutions were formally defined by Bruna et al. \cite{bruna} and elaborated upon by Kipf and Welling in \cite{gcn}, who proposed \textit{graph convolutional networks}. Graph convolutions have also been used in variational graph autoencoding in \cite{vgae}. Each of these approaches are limited to the graph domain, but are relevant to the proposed framework. 

Deep learning on sets has been studied in \cite{pc, deepsets} in the permutation invariant and permutation equivariant cases. Particularly, these architectures employ a linear transformation (adding up the representations) was  between layers of nonlinear transformations to learn continuous representations of permutation invariant objects, which we utilize in this work. Multiset learning has been studied in \cite{multisets} but these methods do not directly employ the context-based inference that can be drawn from the hypergraph structure. Further, methods that make use of membership exclusively, such as PointNet and PointNet++ \cite{pointnet, pointnet++}, and those that make use of the hypergraph structure, such as HGNNs \cite{hgnn}, have been compared, and in the hypergraph context, it is typically better to include contextual information at inference time.

Hypergraph learning is lesser-studied, but a variety of approaches have nonetheless been proposed. In 2007, Zhou et al. proposed methods for hypergraph clustering and embedding in \cite{zhou}, but these methods incur high computational and space complexity. Random walks on hypergraphs have been established, and have likewise been demonstrated as useful in inference tasks \cite{chitra,dwalk,hyperedge2vec}, but these methods do not directly account for the set membership and contextual properties of hyperedges simultaneously and efficiently. Very recently, hypergraph convolution and attention approaches have been proposed \cite{hgnn, sbai} which define a hypergraph Laplacian matrix and perform convolutions on this matrix. Our framework specifically uses a random-walk based model for their own benefits such as parallelizability, scalability, accomodation of inductive learning, and baseline comparisons between random walk models in the graph domain and our own random walk procedure, but convolution approaches could conceivably be integrated into this framework and this line of exploration is left to future work.
\section{Deep Hyperedges}
\begin{figure}[!tbp]
  \centering
  \includegraphics[width=\textwidth]{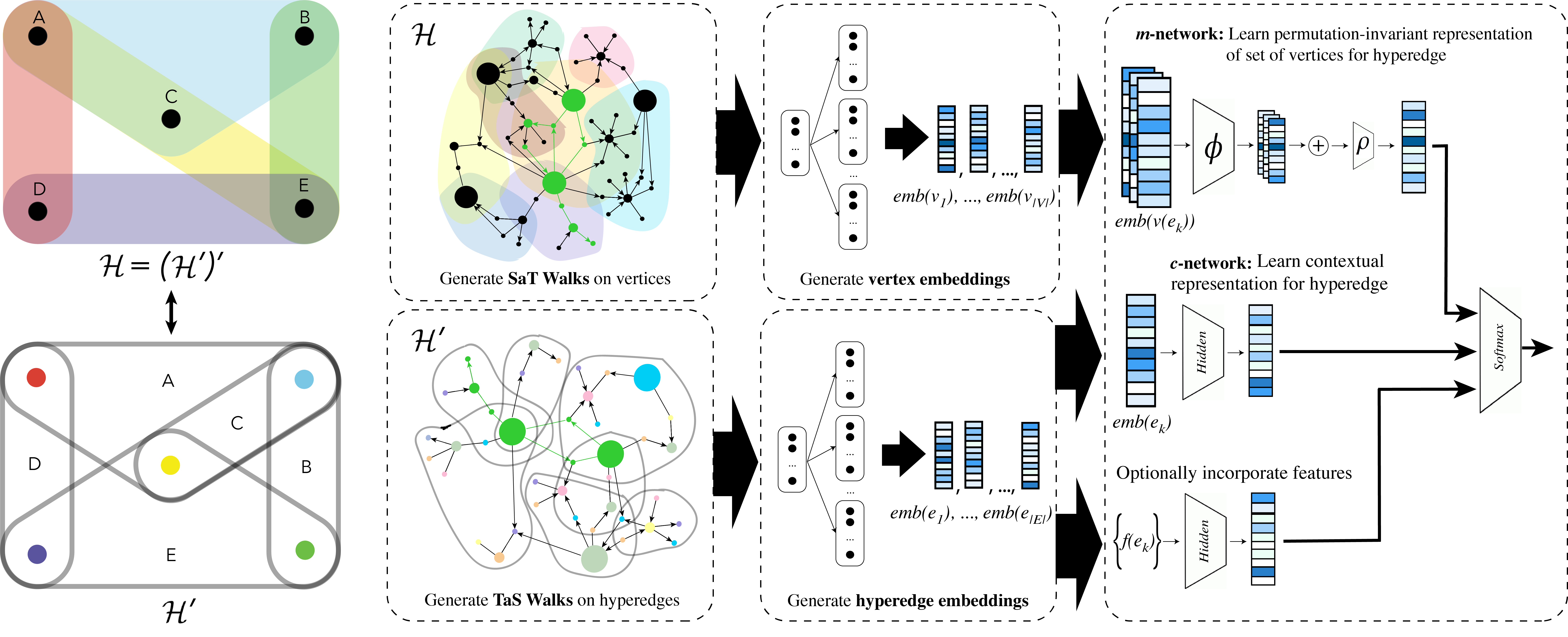}
  \caption{Architectural overview for the Deep Hyperedges model in the transductive setting.}
  \label{fig:DHE_architecture}
\end{figure}

We propose \textit{Deep Hyperedges} (DHE), a framework that utilizes context and set membership to perform transductive and inductive learning on hypergraphs. We will be focusing on classification of hyperedges, but recall that the hyperedges of the dual $\mathcal{H}'$ of a hypergraph $\mathcal{H}$ correspond to the edges of $\mathcal{H}$, so vertex classification may be approached similarly. This is described more in Appendix \ref{app:bg_hgraphs}. While DHE is conducive to random walk, spectral, and convolutional approaches for the vertex and hyperedge embedding steps, we investigate a random walk approach for the sake of parallelizability and scalability \cite{deepwalk} and inductive inference \cite{graphsage}.

\subsection{Random Walks and Embeddings}

To test this framework using random walk models, we proposed a simple new random walk model for hypergraphs, \textit{Subsample and Traverse} (SaT) Walks, which seeks to capture co-member information in each hyperedge. In this procedure, we start at a vertex $v_m$ in a selected hyperedge $e_i$. We define the probability of traversing to be inversely proportional to the cardinality of the current hyperedge; that is, $p = \min(\frac{\alpha}{|e_i|}+\beta, 1)$, where $\alpha, \beta \geq 0$ are tunable parameters. The expectation of samples the walk will draw from a given hyperedge is geometric, and thus influenced by the cardinality of the hyperedge itself. Using SaT Walks, we can construct random walks of vertices in the hypergraph for embedding in the next stage. Likewise, we may perform SaT Walks on $\mathcal{H}'$ to embed the hyperedges of $\mathcal{H}$. For ease of notation, we refer to these walks as \textit{Traverse and Select} (TaS) Walks. Using TaS Walks, we can construct random walks of hyperedges in the hypergraph for contextual embedding in the next stage. One could also easily define \textit{in-out} and \textit{return} parameters for TaS Walks (à la \textit{node2vec} \cite{node2vec}) for controlling topological proximity \textit{vs.} structural similarity (by inclining the search strategy toward BFS or DFS) representations in the embeddings $\Phi(v), \Phi(e)$ of the vertices and hyperedges.
\begin{figure}[!tbp]
  \centering
  \includegraphics[width=\textwidth]{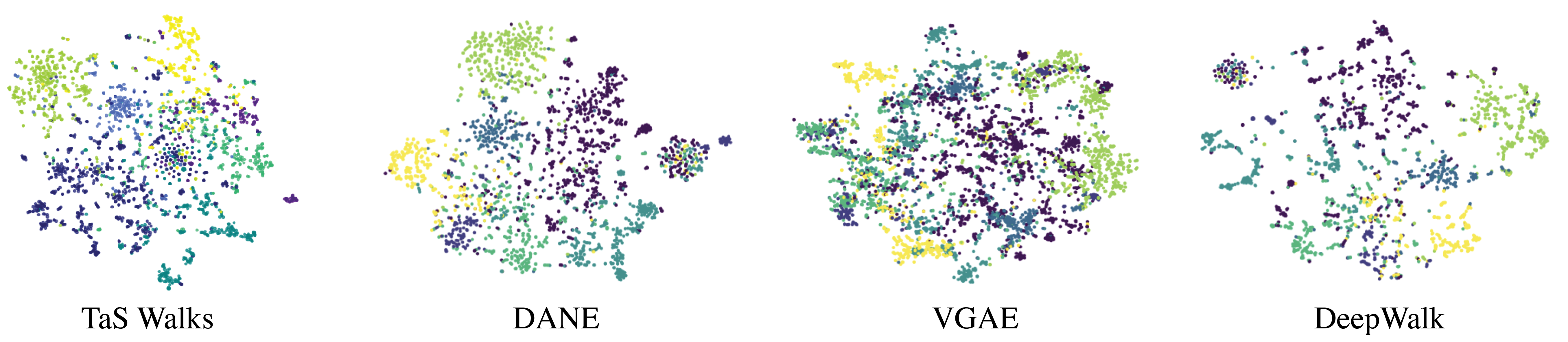}
  \caption{t-SNE \cite{tsne} embedding visualizations comparing the our hyperedge embedding procedure (without additional features), DANE, VGAE, and DeepWalk.}
\end{figure}

\subsection{Contextual and Permutation-Invariant Representation Learning}

At this stage, we employ two distinct networks to address two objectives: 1) we would like to learn a representation of each hyperedge that captures its contextual information, and 2) we'd also like to create a representation of each hyperedge that captures the membership of its vertices in a manner that is invariant to permutation. To address the first objective, we construct a context network ($c$-network) that applies $j$ hidden layers $\textbf{h}$ to the hyperedge embedding $\Phi(e)$ to output a learned contextual representation $\textbf{h}^j(\Phi(e_i)) = c(e_i)$. The second objective requires a linear transformation at a hidden layer for permutation invariance. We use the $\phi$ and $\rho$ networks of the DeepSets architecture. The membership representation network ($m$-network) takes as input $\Phi(v)$ for each $v \in e_i$, apply a nonlinear transformation to these inputs individually (forming our $\phi$ network), add up the representations to obtain $\sum_{v\in e_i}\phi(\Phi(v))$, and finally apply $k$ hidden layers to this representation (forming our $\rho$ network) to obtain a membership representation $m(e_i)$ of the hyperedge $e_i$ that is invariant to permutation. We then concatenate the representations output by the $c$-network and $m$-network and apply $l$ hidden layers and a final softmax layer: $\textup{softmax}(\textbf{h}^l(c(e_i)\mathbin\Vert m(e_i)))$, where $\mathbin\Vert$ denotes concatenation. Optionally, we apply $m$ hidden layers to the input feature vector $f(e_i)$ to obtain $\textbf{h}^m(f(e_i))$, and our formulation would then be \begin{equation}
    \textup{softmax}\left(\textbf{h}^l\left(c(e_i)\mathbin\Vert m(e_i)\mathbin\Vert \textbf{h}^m\left(f(e_i)\right)\right)\right)
\end{equation}

This is the formulation shown in Figure \ref{fig:DHE_architecture}. One could also include a permutation-invariant representations of the features of the vertices of a hyperedge for inclusion using DeepSets, as well. For the inductive formulation, we want to learn a function that can generalize to unseen vertices. Hence, this function will only depend initially on an input feature vector $f(e_i)$ and a uniform sampling and pooling aggregation procedure that arises naturally with SaT Walks and TaS Walks (à la GraphSAGE \cite{graphsage}). The forward propagation algorithm for embedding generation is described in this work, and with these embeddings, we proceed as before.
\section{Evaluation}
\begin{table}
  \caption{Experimental results on the Cora and PubMed datasets.}
  \setlength{\tabcolsep}{2.5pt}
  
  \begin{tabular}{lllllllllllll}
    \toprule
    \multicolumn{1}{c}{}&\multicolumn{6}{c}{Cora} & \multicolumn{6}{c}{PubMed} \\
    \cmidrule(r){2-7}
    \cmidrule(r){8-13}
    \multicolumn{1}{c}{}& \multicolumn{2}{c}{10\%} & \multicolumn{2}{c}{30\%} & \multicolumn{2}{c}{50\%} & \multicolumn{2}{c}{10\%} & \multicolumn{2}{c}{30\%} & \multicolumn{2}{c}{50\%}\\
    \cmidrule(r){2-3}
    \cmidrule(r){4-5}
    \cmidrule(r){6-7}
    \cmidrule(r){8-9}
    \cmidrule(r){10-11}
    \cmidrule(r){12-13}
    Method   & Micro & Macro & Micro & Macro & Micro & Macro & Micro & Macro & Micro & Macro & Micro & Macro \\
    \midrule
    DW & 0.757 & 0.750 & 0.806 & 0.794 & 0.829 & 0.818 & 0.805 & 0.787 & 0.817 & 0.803 & 0.816 & 0.803 \\
    N2V & 0.748 & 0.726 & 0.820 & 0.812 & 0.824 & 0.816 & 0.803 & 0.785 & 0.811 & 0.797 & 0.810 & 0.798 \\
    GraRep & 0.757 & 0.744 & 0.793 & 0.789 & 0.800 & 0.792 & 0.795 & 0.779 & 0.803 & 0.790 & 0.805 & 0.794 \\
    LINE & 0.734 & 0.719 & 0.812 & 0.811 & 0.835 & 0.825 & 0.804 & 0.789 & 0.813 & 0.801 & 0.811 & 0.799 \\
    \cmidrule(r){1-13}
    TADW & 0.751 & 0.723 & 0.801 & 0.780 & 0.835 & 0.825 & 0.836 & 0.834 & 0.859 & 0.858 & 0.864 & 0.863 \\
    ANE & 0.720 & 0.715 & 0.803 & 0.791 & 0.812 & 0.799 & 0.798 & 0.788 & 0.826 & 0.819 & 0.828 & 0.820 \\
    GAE & 0.769 & 0.757 & 0.806 & 0.792 & 0.810 & 0.799 & 0.829 & 0.824 & 0.831 & 0.826 & 0.831 & 0.826 \\
    VGAE & 0.789 & 0.774 & 0.805 & 0.791 & 0.812 & 0.799 & 0.830 & 0.824 & 0.835 & 0.829 & 0.836 & 0.830 \\
    SAGE & 0.763 & 0.748 & 0.810 & 0.800 & 0.815 & 0.808 & 0.817 & 0.809 & 0.825 & 0.816 & 0.827 & 0.818 \\
    DANE & 0.787 & \textbf{0.775} & 0.828 & 0.813 & \textbf{0.850} & \textbf{0.838} & 0.861 & 0.858 & 0.873 & 0.871 & 0.878 & 0.875 \\
    \cmidrule(r){1-13}
    Ours & \textbf{0.790} & 0.771 & \textbf{0.830} & \textbf{0.818} & 0.842 & 0.833 & \textbf{0.870} & \textbf{0.868} & \textbf{0.887} & \textbf{0.886} & \textbf{0.894} & \textbf{0.891} \\
    \bottomrule
  \end{tabular}
\end{table}

To evaluate DHE on benchmark citation network datasets that have been configured as hypergraphs where $e_i$ is paper $i$ and its 1-neighborhood (described in Appendix \ref{app:benchmark}), we compare to several state-of-the-art approaches (in order): DeepWalk (DW) by Perozzi et al., 2014 \cite{deepwalk}, \textit{node2vec} (N2V) by Grover et al., 2016 \cite{node2vec}, GraRep by Cao et al., 2015 \cite{grarep}, and LINE by Tang et al., 2015 \cite{LINE}, which do not use network attributes; and TADW by Yang et al., 2015 \cite{tadw}, ANE by Huang et al., \cite{ane}, Graph Auto-Encoder (GAE) by Kipf and Welling, 2016 \cite{vgae}, Variational Graph Auto-Encoder (VGAE) by Kipf and Welling, 2016 \cite{vgae}, GraphSAGE (SAGE) by Hamilton et al., 2017 \cite{graphsage}, and DANE by Gao and Huang, 2018 \cite{DANE}, which do use network attributes, as our model does. We benchmark $F_1$ scores for both datasets with randomly sampled train:test splits of 10:90, 30:70, and 50:50, as reported by \cite{DANE} (none is used for validation). We also explore model performance on several nonstandard hypergraph datasets in Appendix \ref{app:other_datasets}.

We use 25 random walks of length 25 for each hyperedge and vertex and embed each hyperedge and vertex into $\mathbb{R}^{128}$ and $\mathbb{R}^{16}$, respectively, using skip-gram. Each hidden layer in each of our networks has 100 neurons, with the exception of of the output of the $c$-network, which has 30 neurons. We use dropout \cite{dropout} in most of the hidden layers to prevent overfitting, and ReLU for the activation function for each layer except for the output (softmax) and $\phi$ network layers ($\tanh$, as suggested in \cite{deepsets}). Finally, we use SGD as our optimization algorithm and categorical cross-entropy loss. The scores reported are the average of 5 runs for each test. Our model scales well and has been tested on hyperedges of cardinalities of up to $1000$ while maintaining low epoch runtime (less than $30$ seconds per epoch on a CPU). DHE outperforms the other methods in most cases, with the additional, unique advantage that it can be extended to hypergraphs.

\section{Conclusion}
In this work, we've proposed Deep Hyperedges (DHE), a framework that jointly uses contextual and permutation-invariant vertex membership properties of hyperedges in hypergraphs to perform classification and regression in transductive and inductive learning settings. We've also proposed a random walk model that can be used for obtaining embeddings of hyperedges and vertices for evaluation. With these, demonstrated that DHE achieves and oftentimes surpasses state-of-the-art performance on benchmark graph datasets and explored nonstandard hypergraph datasets in Appendix \ref{app:other_datasets}). We've also identified several exciting avenues of future work, including deeper exploration into the inductive learning case and usage of convolutional models in the embedding step. One aim of this framework is to provide flexibility in integrating of other approaches, and in doing so, encourage collaborative efforts towards cracking the hypergraph learning problem.
\bibliographystyle{IEEEtran}
\bibliography{IEEEabrv,references}
\begin{appendix}\label{appendix}
\section{Supplementary: Background on Graphs and Hypergraphs}\label{app:bg_hgraphs}
A \textit{hypergraph} $\mathcal{H}$ = $(V,E)$ is comprised of a finite set of vertices $V = \{v_1, v_2, \ldots, v_n\}$ and a set of hyperedges $E = \{e_1, e_2, \ldots, e_m\} \subseteq 2^V$. We consider connected hypergraphs with $|V| \geq 2$. The \textit{dual}, $\mathcal{H}'$, of a hypergraph $\mathcal{H}$ is a hypergraph such that the hyperedges and vertices of $\mathcal{H}'$ are given by the vertices and hyperedges of $\mathcal{H}$, respectively. That is, a vertex $v'_i$ of $\mathcal{H}'$ is a member of a hyperedge $e'_j$ of $\mathcal{H}'$ if and only if its corresponding hyperedge $e_i$ of $\mathcal{H}$ contained the vertex $v_j$ of $\mathcal{H}$ corresponding to $e'_j$. Note that $(\mathcal{H}')' = \mathcal{H}$. A \textit{graph} $\mathcal{G}$ is a hypergraph where $|e_i| = 2$ for each $e_i\in E$. Graphs are well-studied in the field of machine learning, but are not capable completely representing the information captured in hypergraphs generally. Ihler et al. showed in \cite{ihler}, for instance, that if $|V|\geq 4$, there does not exist a representation of a hypergraph by a graph with the same cut properties in the general case. This fact has practical implications as well: for instance, one cannot represent the protein complexes in the CORUM dataset \cite{corum} using pairwise relationships, as the proteins in a complex may not interact with each other independently. These hyperedges are said to be \textit{indecomposable} \cite{dhne}. Hence, while the theory built for studying graphs can be utilized to some capacity in the hypergraph context, there is a need for learning procedures that can effectively generalize to hypergraphs.

\section{Supplementary: Background on Transductive and Inductive Learning}\label{app:bg_learning}
In \textit{transductive} (or \textit{semi-supervised}) inference tasks, one often seeks to learn from a small amount of labeled training data, where the model has access to labeled and unlabeled data at training time. Formally, we have training instances $\{\textbf{x}_i\}_1^n$ where $\{\textbf{x}_i\}_1^l$ are labeled instances and $\{\textbf{x}_i\}_{l+1}^{l+u}$ unlabeled instances, and corresponding labels $y_i$ in $\{y_i\}_1^n$. Our aim is to learn a function $F : \{\textbf{x}_i\}_1^n \rightarrow \{y_i\}_1^n; \ \textbf{x}_i \mapsto y_i$. Typically, in the case of transductive learning on graphs and hypergraphs, we seek to leverage topological information to represent the vertices in some continuous vector space $\mathbb{R}^{d}$ by embeddings which capture the vertices' or hyperedges' context (homophily and structural equivalence). As such, in the pre-training procedure for finding vertex embeddings, we want to find an embedding function $\Phi : V \rightarrow \mathbb{R}^d$ that maximizes the likelihood of observing a vertex in the sampled neighborhood $N(v)$ of $v$ given $\Phi(v)$: \begin{equation}
 \max_\Phi \displaystyle\sum_{v\in V}\log \mathbb{P}(N(v) \ | \ \Phi(v))
\end{equation}
The procedure for finding contextual embeddings for hyperedges is similar. Once we've learned $\Phi$, we can use the embeddings $\Phi(v)$ to learn $F$ in our transductive learning procedure.

In \textit{inductive} (or \textit{supervised}) inference tasks, we are given a training sample $\{\textbf{x}_i\}_1^n \subseteq X$ to be seen by our model, and we want to learn a function $g : X \rightarrow Y; \ \textbf{x}_i \mapsto y_i$ that can generalize to unseen instances. This type of learning is particularly useful for dynamic graphs and hypergraphs, when we may find unseen vertices as time passes, or when we want to apply transfer learning to new graphs and hypergraphs altogether. Here, the representation learning function $\Phi$ is typically dependent on the input features $f(v)$ for a vertex $v$.

Several approaches have been proposed for each of these learning paradigms. Each has its uses, and each can be applied using this framework. However, for testing purposes, we focus primarily on the transductive case, leaving a quantitative investigation of the inductive case in this framework to a future work.

\section{Supplementary: Benchmark Dataset Descriptions}\label{app:benchmark}
Because far more literature exists on graph deep learning than does on hypergraph deep learning, a typical benchmarking approach is to take a graph dataset, represent the data in a way that creates hyperedges with cardinality greater than two (without modifying the data itself), and compare with graph-based learning benchmarks. For this phase of the evaluation, we used the Cora and PubMed datasets, citation networks in which vertices are papers, edges are citations, and each paper is labeled as being one of 7 or 3 fairly balanced classes, respectively. Each vertex also has a feature vector, which is a publication content representation in $\mathbb{F}_2^{1433}$ and $\mathbb{R}^{500}$ for the Cora and PubMed datasets, respectively. The hypergraph extension involves creating a hyperedge for each paper where the paper is the centroid and its 1-neighborhood forms the hyperedge. As the mapping between papers and hyperedges is bijective, it remains to classify the hyperedge for the inference task.

\section{Supplementary: Non-Standard Datasets and Comparisons Between Models}\label{app:other_datasets}
In this appendix, we explore and compare performance results between the three models presented (Deep Hyperedges, DeepSets + SaT Walks, Multilayer Perceptron + TaS Walks), as well as between the seven datasets we investigated. The training:validation:test split used is 80:10:10, and network features are not used---only the hypergraph structure. As mentioned, these results could improve in a future work with the use of graph convolutions in this framework, as well as model tweaking. 

From left to right, we can view a t-SNE plot of the hyperedge embeddings constructed using TaS Walks, training accuracy, training loss, validation accuracy, and validation loss. The green plot represents DHE, the blue plot represents DeepSets + SaT Walks, and the red plot represents MLP + TaS Walks. This is meant to provide an overview at-a-glance of the relative effectiveness and trends of each model for each dataset, and across datasets. We observe that DHE typically performs the best in validation and test accuracy, and that the relative level of performance between DeepSets + SaT Walks and MLP + TaS Walks depends on how important cardinality, vertex membership, and context are in the data.

These tests are all available as Jupyter Notebooks for ease of access and convenience of testing on other hypergraph data.\footnote{Code: https://github.com/Josh-Payne/deep-hyperedges}

\includegraphics[width=\textwidth]{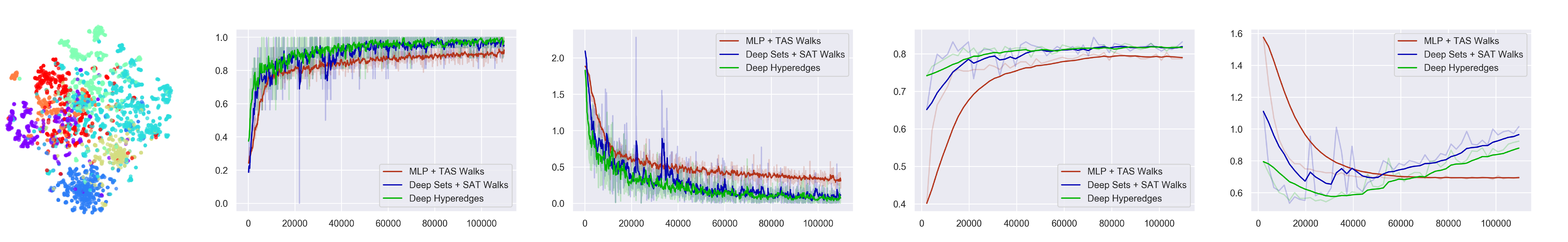}
The \textbf{Cora} dataset\footnote{https://relational.fit.cvut.cz/dataset/CORA} is a computer science publication citation network dataset where vertices are papers and hyperedges are the 1-neighborhood of a centroid paper. Each paper (and thus each hyperedge) is classified into one of seven classes based on topic. DHE achieved around 82.29\% test accuracy.

\includegraphics[width=\textwidth]{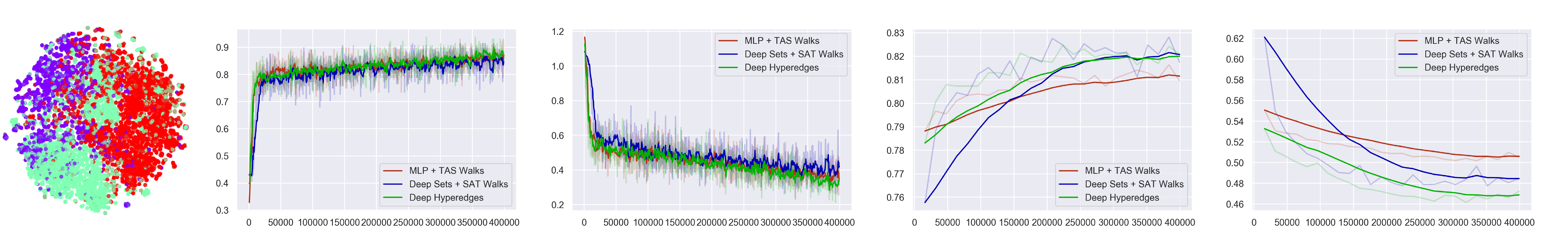}
The \textbf{PubMed} diabetes publication citation network dataset\footnote{https://linqs.soe.ucsc.edu/data} is a network where vertices are papers and hyperedges are the cited works of a given paper. Each paper is classified into one of three classes based on the type of diabetes it studies. DHE achieved around 82.35\% test accuracy.

\includegraphics[width=\textwidth]{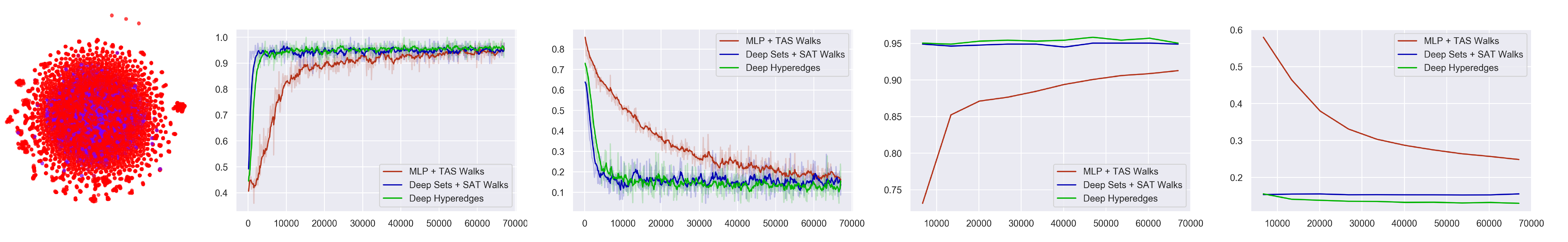}
The \textbf{CORUM} protein complex dataset\footnote{https://mips.helmholtz-muenchen.de/corum/} represents a network where vertices are proteins and hyperedges are collections of proteins that interact to form a protein complex. Each hyperedge is labeled based on whether or not the collection forms a protein complex. We generate negative examples by selecting a $n$ proteins at random to form a negative hyperedge, where $n$ is selected uniformly between the minimum hyperedge cardinality and the maximum hyperedge cardinality. DHE achieves 95.53\% test accuracy, with DeepSets + SaT Walks close behind, as a critical signal factor here is cardinality.

\includegraphics[width=\textwidth]{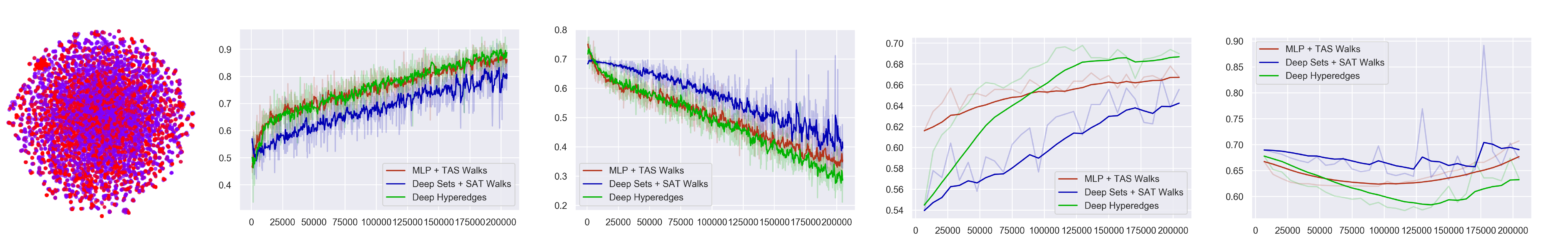}
\textbf{CORUM} dataset, same distribution. We generate negative examples by selecting a $n$ proteins at random to form a negative hyperedge, where $n$ is selected from the distribution in the data. This is a trickier dataset, and DHE achieves 70.23\% test accuracy. The key insight brought by differences in cardinality enjoyed by DeepSets + SaT Walks in the previous dataset are no more, as it trails behind MLP + TaS Walks. Perhaps if we'd let it train for more epochs, it would've caught up, as DeepSets + SaT Walks captures context, as well---just not as readily.

\includegraphics[width=\textwidth]{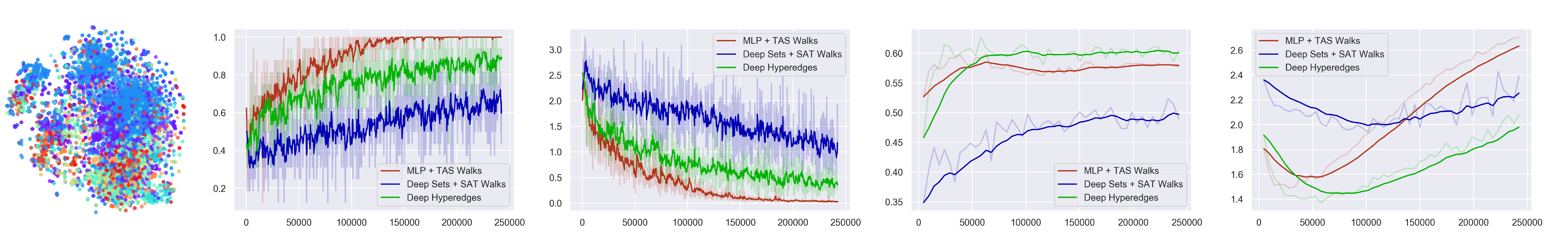}
The \textbf{Meetups} dataset\footnote{https://www.kaggle.com/sirpunch/meetups-data-from-meetupcom} is a highly unbalanced social networking dataset where vertices are members and hyperedges are meetup events. Each meetup event is classified into one of 33 types. The model performed fairly decently, achieving around 60.10\% accuracy.

\includegraphics[width=\textwidth]{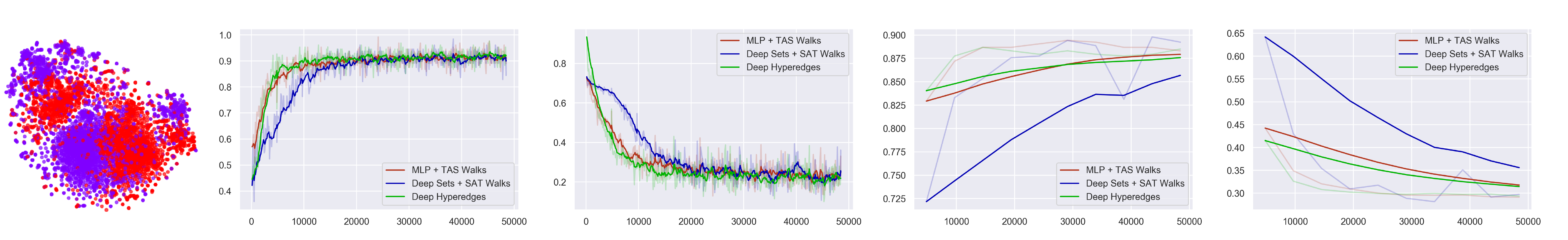}
\textbf{Meetups} dataset, balanced. We grouped ``Tech" and ``Career \& Business" meetups together into one class and every other type of meetup into another class to give the dataset more balance. This time, DHE performed much better, achieving around 88.31\% test accuracy. DeepSets + SaT Walks performed even better, with 91.31\% test accuracy.

\includegraphics[width=\textwidth]{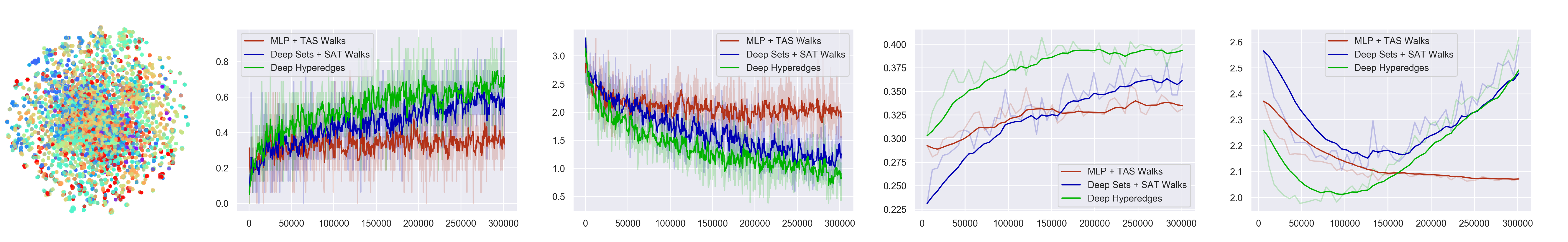}
The \textbf{DisGeNet}\footnote{http://www.disgenet.org/downloads} dataset is a fairly unbalanced dataset with 23 classes. It is a disease genomics dataset where vertices are genes and hyperedges are diseases. Each disease is classified into one of 23 MeSH codes (if it has multiple, we randomly select one). The model achieved around 40\% test accuracy---more than we expected, given that additional features are not being trained on, and the linear inseparability of the hyperedge embeddings.

\end{appendix}

\end{document}